\documentclass{edm_template}
\usepackage{booktabs}
\usepackage{tabularx}
\usepackage{multirow}
\usepackage{hhline}
\usepackage{color}
\usepackage[dvipsnames]{xcolor}
\usepackage{url}
\usepackage{amsmath,epsfig} 
\usepackage{amsfonts}
\usepackage{amssymb}
\usepackage{amsmath,epsfig} 
\usepackage{subfigure} 
\usepackage{paralist}
\usepackage{booktabs}
\usepackage[draft]{hyperref}
\usepackage{color}
\usepackage{microtype}
\DeclareGraphicsExtensions{.pdf,.jpeg,.png,.epbibdesks}
\usepackage[T1]{fontenc}
\usepackage{epstopdf}
\usepackage{graphicx}

\usepackage{floatrow}
\newfloatcommand{capbtabbox}{table}[][\FBwidth]
\usepackage[ruled,vlined]{algorithm2e}
\SetArgSty{textup}
\usepackage{yhmath}
\allowdisplaybreaks[4]
\usepackage{balance}

\usepackage{balance}  
\usepackage{mdframed}
\usepackage{subfigure} 
\usepackage{amssymb}
\usepackage{amsfonts}
\usepackage{mathrsfs}
\usepackage{xspace}
\usepackage{bm}
\usepackage{upgreek}

\newcommand{\safemath}[2]{\newcommand{#1}{\ensuremath{#2}\xspace}}

\safemath{\bma}{\mathbf{a}}
\safemath{\bmb}{\mathbf{b}}
\safemath{\bmc}{\mathbf{c}}
\safemath{\bmd}{\mathbf{d}}
\safemath{\bme}{\mathbf{e}}
\safemath{\bmf}{\mathbf{f}}
\safemath{\bmg}{\mathbf{g}}
\safemath{\bmh}{\mathbf{h}}
\safemath{\bmi}{\mathbf{i}}
\safemath{\bmj}{\mathbf{j}}
\safemath{\bmk}{\mathbf{k}}
\safemath{\bml}{\mathbf{l}}
\safemath{\bmm}{\mathbf{m}}
\safemath{\bmn}{\mathbf{n}}
\safemath{\bmo}{\mathbf{o}}
\safemath{\bmp}{\mathbf{p}}
\safemath{\bmq}{\mathbf{q}}
\safemath{\bmr}{\mathbf{r}}
\safemath{\bms}{\mathbf{s}}
\safemath{\bmt}{\mathbf{t}}
\safemath{\bmu}{\mathbf{u}}
\safemath{\bmv}{\mathbf{v}}
\safemath{\bmw}{\mathbf{w}}
\safemath{\bmx}{\mathbf{x}}
\safemath{\bmy}{\mathbf{y}}
\safemath{\bmz}{\mathbf{z}}
\safemath{\bmzero}{\mathbf{0}}
\safemath{\bmone}{\mathbf{1}}

\bmdefine{\biad}{a}
\bmdefine{\bibd}{b}
\bmdefine{\bicd}{c}
\bmdefine{\bidd}{d}
\bmdefine{\bied}{e}
\bmdefine{\bifd}{f}
\bmdefine{\bigd}{g}
\bmdefine{\bihd}{h}
\bmdefine{\biid}{i}
\bmdefine{\bijd}{j}
\bmdefine{\bikd}{k}
\bmdefine{\bild}{l}
\bmdefine{\bimd}{m}
\bmdefine{\bind}{n}
\bmdefine{\biod}{o}
\bmdefine{\bipd}{p}
\bmdefine{\biqd}{q}
\bmdefine{\bird}{r}
\bmdefine{\bisd}{s}
\bmdefine{\bitd}{t}
\bmdefine{\biud}{u}
\bmdefine{\bivd}{v}
\bmdefine{\biwd}{w}
\bmdefine{\bixd}{x}
\bmdefine{\biyd}{y}
\bmdefine{\bizd}{z}

\bmdefine{\bixid}{\xi}
\bmdefine{\bilambdad}{\lambda}
\bmdefine{\bimud}{\mu}
\bmdefine{\bithetad}{\theta}
\bmdefine{\biphid}{\phi}
\bmdefine{\bideltad}{\delta}

\safemath{\bmia}{\biad}
\safemath{\bmib}{\bibd}
\safemath{\bmic}{\bicd}
\safemath{\bmid}{\bidd}
\safemath{\bmie}{\bied}
\safemath{\bmif}{\bifd}
\safemath{\bmig}{\bigd}
\safemath{\bmih}{\bihd}
\safemath{\bmii}{\biid}
\safemath{\bmij}{\bijd}
\safemath{\bmik}{\bikd}
\safemath{\bmil}{\bild}
\safemath{\bmim}{\bimd}
\safemath{\bmin}{\bind}
\safemath{\bmio}{\biod}
\safemath{\bmip}{\bipd}
\safemath{\bmiq}{\biqd}
\safemath{\bmir}{\bird}
\safemath{\bmis}{\bisd}
\safemath{\bmit}{\bitd}
\safemath{\bmiu}{\biud}
\safemath{\bmiv}{\bivd}
\safemath{\bmiw}{\biwd}
\safemath{\bmix}{\bixd}
\safemath{\bmiy}{\biyd}
\safemath{\bmiz}{\bizd}

\safemath{\bmxi}{\bixid}
\safemath{\bmlambda}{\bilambdad}
\safemath{\bmmu}{\bimud}
\safemath{\bmtheta}{\bithetad}
\safemath{\bmphi}{\biphid}
\safemath{\bmdelta}{\bideltad}

\safemath{\bA}{\mathbf{A}}
\safemath{\bB}{\mathbf{B}}
\safemath{\bC}{\mathbf{C}}
\safemath{\bD}{\mathbf{D}}
\safemath{\bE}{\mathbf{E}}
\safemath{\bF}{\mathbf{F}}
\safemath{\bG}{\mathbf{G}}
\safemath{\bH}{\mathbf{H}}
\safemath{\bI}{\mathbf{I}}
\safemath{\bJ}{\mathbf{J}}
\safemath{\bK}{\mathbf{K}}
\safemath{\bL}{\mathbf{L}}
\safemath{\bM}{\mathbf{M}}
\safemath{\bN}{\mathbf{N}}
\safemath{\bO}{\mathbf{O}}
\safemath{\bP}{\mathbf{P}}
\safemath{\bQ}{\mathbf{Q}}
\safemath{\bR}{\mathbf{R}}
\safemath{\bS}{\mathbf{S}}
\safemath{\bT}{\mathbf{T}}
\safemath{\bU}{\mathbf{U}}
\safemath{\bV}{\mathbf{V}}
\safemath{\bW}{\mathbf{W}}
\safemath{\bX}{\mathbf{X}}
\safemath{\bY}{\mathbf{Y}}
\safemath{\bZ}{\mathbf{Z}}

\safemath{\bZero}{\mathbf{0}}
\safemath{\bOne}{\mathbf{1}}
\safemath{\bDelta}{\mathbf{\Delta}}
\safemath{\bLambda}{\mathbf{\UpLambda}}
\safemath{\bPhi}{\mathbf{\Upphi}}
\safemath{\bSigma}{\mathbf{\Upsigma}}
\safemath{\bOmega}{\mathbf{\Upomega}}
\safemath{\bTheta}{\mathbf{\Uptheta}}

\bmdefine{\biAd}{A}
\bmdefine{\biBd}{B}
\bmdefine{\biCd}{C}
\bmdefine{\biDd}{D}
\bmdefine{\biEd}{E}
\bmdefine{\biFd}{F}
\bmdefine{\biGd}{G}
\bmdefine{\biHd}{H}
\bmdefine{\biId}{I}
\bmdefine{\biJd}{J}
\bmdefine{\biKd}{K}
\bmdefine{\biLd}{L}
\bmdefine{\biMd}{M}
\bmdefine{\biNd}{N}
\bmdefine{\biOd}{O}
\bmdefine{\biPd}{P}
\bmdefine{\biQd}{Q}
\bmdefine{\biRd}{R}
\bmdefine{\biSd}{S}
\bmdefine{\biTd}{T}
\bmdefine{\biUd}{U}
\bmdefine{\biVd}{V}
\bmdefine{\biWd}{W}
\bmdefine{\biXd}{X}
\bmdefine{\biYd}{Y}
\bmdefine{\biZd}{Z}

\bmdefine{\biDelta}{\Delta}
\bmdefine{\biLambda}{\Lambda}
\bmdefine{\biPhi}{\Phi}
\bmdefine{\biSigma}{\Sigma}
\bmdefine{\biOmega}{\Omega}
\bmdefine{\biTheta}{\Theta}

\safemath{\bimA}{\biAd}
\safemath{\bimB}{\biBd}
\safemath{\bimC}{\biCd}
\safemath{\bimD}{\biDd}
\safemath{\bimE}{\biEd}
\safemath{\bimF}{\biFd}
\safemath{\bimG}{\biGd}
\safemath{\bimH}{\biHd}
\safemath{\bimI}{\biId}
\safemath{\bimJ}{\biJd}
\safemath{\bimK}{\biKd}
\safemath{\bimL}{\biLd}
\safemath{\bimM}{\biMd}
\safemath{\bimN}{\biNd}
\safemath{\bimO}{\biOd}
\safemath{\bimP}{\biPd}
\safemath{\bimQ}{\biQd}
\safemath{\bimR}{\biRd}
\safemath{\bimS}{\biSd}
\safemath{\bimT}{\biTd}
\safemath{\bimU}{\biUd}
\safemath{\bimV}{\biVd}
\safemath{\bimW}{\biWd}
\safemath{\bimX}{\biXd}
\safemath{\bimY}{\biYd}
\safemath{\bimZ}{\biZd}

\safemath{\bimDelta}{\biDelta}
\safemath{\bimLambda}{\biLambda}
\safemath{\bimPhi}{\biPhi}
\safemath{\bimSigma}{\biSigma}
\safemath{\bimOmega}{\biOmega}
\safemath{\bimTheta}{\biTheta}

\safemath{\setA}{\mathcal{A}}
\safemath{\setB}{\mathcal{B}}
\safemath{\setC}{\mathcal{C}}
\safemath{\setD}{\mathcal{D}}
\safemath{\setE}{\mathcal{E}}
\safemath{\setF}{\mathcal{F}}
\safemath{\setG}{\mathcal{G}}
\safemath{\setH}{\mathcal{H}}
\safemath{\setI}{\mathcal{I}}
\safemath{\setJ}{\mathcal{J}}
\safemath{\setK}{\mathcal{K}}
\safemath{\setL}{\mathcal{L}}
\safemath{\setM}{\mathcal{M}}
\safemath{\setN}{\mathcal{N}}
\safemath{\setO}{\mathcal{O}}
\safemath{\setP}{\mathcal{P}}
\safemath{\setQ}{\mathcal{Q}}
\safemath{\setR}{\mathcal{R}}
\safemath{\setS}{\mathcal{S}}
\safemath{\setT}{\mathcal{T}}
\safemath{\setU}{\mathcal{U}}
\safemath{\setV}{\mathcal{V}}
\safemath{\setW}{\mathcal{W}}
\safemath{\setX}{\mathcal{X}}
\safemath{\setY}{\mathcal{Y}}
\safemath{\setZ}{\mathcal{Z}}
\safemath{\emptySet}{\varnothing}

\safemath{\colA}{\mathscr{A}}
\safemath{\colB}{\mathscr{B}}
\safemath{\colC}{\mathscr{C}}
\safemath{\colD}{\mathscr{D}}
\safemath{\colE}{\mathscr{E}}
\safemath{\colF}{\mathscr{F}}
\safemath{\colG}{\mathscr{G}}
\safemath{\colH}{\mathscr{H}}
\safemath{\colI}{\mathscr{I}}
\safemath{\colJ}{\mathscr{J}}
\safemath{\colK}{\mathscr{K}}
\safemath{\colL}{\mathscr{L}}
\safemath{\colM}{\mathscr{M}}
\safemath{\colN}{\mathscr{N}}
\safemath{\colO}{\mathscr{O}}
\safemath{\colP}{\mathscr{P}}
\safemath{\colQ}{\mathscr{Q}}
\safemath{\colR}{\mathscr{R}}
\safemath{\colS}{\mathscr{S}}
\safemath{\colT}{\mathscr{T}}
\safemath{\colU}{\mathscr{U}}
\safemath{\colV}{\mathscr{V}}
\safemath{\colW}{\mathscr{W}}
\safemath{\colX}{\mathscr{X}}
\safemath{\colY}{\mathscr{Y}}
\safemath{\colZ}{\mathscr{Z}}

\safemath{\opA}{\mathbb{A}}
\safemath{\opB}{\mathbb{B}}
\safemath{\opC}{\mathbb{C}}
\safemath{\opD}{\mathbb{D}}
\safemath{\opE}{\mathbb{E}}
\safemath{\opF}{\mathbb{F}}
\safemath{\opG}{\mathbb{G}}
\safemath{\opH}{\mathbb{H}}
\safemath{\opI}{\mathbb{I}}
\safemath{\opJ}{\mathbb{J}}
\safemath{\opK}{\mathbb{K}}
\safemath{\opL}{\mathbb{L}}
\safemath{\opM}{\mathbb{M}}
\safemath{\opN}{\mathbb{N}}
\safemath{\opO}{\mathbb{O}}
\safemath{\opP}{\mathbb{P}}
\safemath{\opQ}{\mathbb{Q}}
\safemath{\opR}{\mathbb{R}}
\safemath{\opS}{\mathbb{S}}
\safemath{\opT}{\mathbb{T}}
\safemath{\opU}{\mathbb{U}}
\safemath{\opV}{\mathbb{V}}
\safemath{\opW}{\mathbb{W}}
\safemath{\opX}{\mathbb{X}}
\safemath{\opY}{\mathbb{Y}}
\safemath{\opZ}{\mathbb{Z}}
\safemath{\opZero}{\mathbb{O}}
\safemath{\identityop}{\opI}

\safemath{\veca}{\bma}
\safemath{\vecb}{\bmb}
\safemath{\vecc}{\bmc}
\safemath{\vecd}{\bmd}
\safemath{\vece}{\bme}
\safemath{\vecf}{\bmf}
\safemath{\vecg}{\bmg}
\safemath{\vech}{\bmh}
\safemath{\veci}{\bmi}
\safemath{\vecj}{\bmj}
\safemath{\veck}{\bmk}
\safemath{\vecl}{\bml}
\safemath{\vecm}{\bmm}
\safemath{\vecn}{\bmn}
\safemath{\veco}{\bmo}
\safemath{\vecp}{\bmp}
\safemath{\vecq}{\bmq}
\safemath{\vecr}{\bmr}
\safemath{\vecs}{\bms}
\safemath{\vect}{\bmt}
\safemath{\vecu}{\bmu}
\safemath{\vecv}{\bmv}
\safemath{\vecw}{\bmw}
\safemath{\vecx}{\bmx}
\safemath{\vecy}{\bmy}
\safemath{\vecz}{\bmz}

\safemath{\veczero}{\bmzero}
\safemath{\vecone}{\bmone}
\safemath{\vecxi}{\bmxi}
\safemath{\veclambda}{\bmlambda}
\safemath{\vecmu}{\bmmu}
\safemath{\vectheta}{\bmtheta}
\safemath{\vecphi}{\bmphi}
\safemath{\vecdelta}{\bmdelta}

\safemath{\matA}{\bA}
\safemath{\matB}{\bB}
\safemath{\matC}{\bC}
\safemath{\matD}{\bD}
\safemath{\matE}{\bE}
\safemath{\matF}{\bF}
\safemath{\matG}{\bG}
\safemath{\matH}{\bH}
\safemath{\matI}{\bI}
\safemath{\matJ}{\bJ}
\safemath{\matK}{\bK}
\safemath{\matL}{\bL}
\safemath{\matM}{\bM}
\safemath{\matN}{\bN}
\safemath{\matO}{\bO}
\safemath{\matP}{\bP}
\safemath{\matQ}{\bQ}
\safemath{\matR}{\bR}
\safemath{\matS}{\bS}
\safemath{\matT}{\bT}
\safemath{\matU}{\bU}
\safemath{\matV}{\bV}
\safemath{\matW}{\bW}
\safemath{\matX}{\bX}
\safemath{\matY}{\bY}
\safemath{\matZ}{\bZ}
\safemath{\matzero}{\bmzero}

\safemath{\matDelta}{\bDelta}
\safemath{\matLambda}{\bLambda}
\safemath{\matPhi}{\bPhi}
\safemath{\matSigma}{\bSigma}
\safemath{\matOmega}{\bOmega}
\safemath{\matTheta}{\bTheta}

\safemath{\matidentity}{\matI}
\safemath{\matone}{\matO}

\safemath{\rnda}{A}
\safemath{\rndb}{B}
\safemath{\rndc}{C}
\safemath{\rndd}{D}
\safemath{\rnde}{E}
\safemath{\rndf}{F}
\safemath{\rndg}{G}
\safemath{\rndh}{H}
\safemath{\rndi}{I}
\safemath{\rndj}{J}
\safemath{\rndk}{K}
\safemath{\rndl}{L}
\safemath{\rndm}{M}
\safemath{\rndn}{N}
\safemath{\rndo}{O}
\safemath{\rndp}{P}
\safemath{\rndq}{Q}
\safemath{\rndr}{R}
\safemath{\rnds}{S}
\safemath{\rndt}{T}
\safemath{\rndu}{U}
\safemath{\rndv}{V}
\safemath{\rndw}{W}
\safemath{\rndx}{X}
\safemath{\rndy}{Y}
\safemath{\rndz}{Z}

\safemath{\rveca}{\bimA}
\safemath{\rvecb}{\bimB}
\safemath{\rvecc}{\bimC}
\safemath{\rvecd}{\bimD}
\safemath{\rvece}{\bimE}
\safemath{\rvecf}{\bimF}
\safemath{\rvecg}{\bimG}
\safemath{\rvech}{\bimH}
\safemath{\rveci}{\bimI}
\safemath{\rvecj}{\bimJ}
\safemath{\rveck}{\bimK}
\safemath{\rvecl}{\bimL}
\safemath{\rvecm}{\bimM}
\safemath{\rvecn}{\bimN}
\safemath{\rveco}{\bomO}
\safemath{\rvecp}{\bimP}
\safemath{\rvecq}{\bimQ}
\safemath{\rvecr}{\bimR}
\safemath{\rvecs}{\bimS}
\safemath{\rvect}{\bimT}
\safemath{\rvecu}{\bimU}
\safemath{\rvecv}{\bimV}
\safemath{\rvecw}{\bimW}
\safemath{\rvecx}{\bimX}
\safemath{\rvecy}{\bimY}
\safemath{\rvecz}{\bimZ}

\safemath{\rvecxi}{\bmxi}
\safemath{\rveclambda}{\bmlambda}
\safemath{\rvecmu}{\bmmu}
\safemath{\rvectheta}{\bmtheta}
\safemath{\rvecphi}{\bmphi}

\safemath{\rmatA}{\bimA}
\safemath{\rmatB}{\bimB}
\safemath{\rmatC}{\bimC}
\safemath{\rmatD}{\bimD}
\safemath{\rmatE}{\bimE}
\safemath{\rmatF}{\bimF}
\safemath{\rmatG}{\bimG}
\safemath{\rmatH}{\bimH}
\safemath{\rmatI}{\bimI}
\safemath{\rmatJ}{\bimJ}
\safemath{\rmatK}{\bimK}
\safemath{\rmatL}{\bimL}
\safemath{\rmatM}{\bimM}
\safemath{\rmatN}{\bimN}
\safemath{\rmatO}{\bimO}
\safemath{\rmatP}{\bimP}
\safemath{\rmatQ}{\bimQ}
\safemath{\rmatR}{\bimR}
\safemath{\rmatS}{\bimS}
\safemath{\rmatT}{\bimT}
\safemath{\rmatU}{\bimU}
\safemath{\rmatV}{\bimV}
\safemath{\rmatW}{\bimW}
\safemath{\rmatX}{\bimX}
\safemath{\rmatY}{\bimY}
\safemath{\rmatZ}{\bimZ}

\safemath{\rmatDelta}{\bimDelta}
\safemath{\rmatLambda}{\bimLambda}
\safemath{\rmatPhi}{\bimPhi}
\safemath{\rmatSigma}{\bimSigma}
\safemath{\rmatOmega}{\bimOmega}
\safemath{\rmatTheta}{\bimTheta}

\usepackage{amssymb}
\usepackage{amsfonts}
\usepackage{mathrsfs}
\usepackage{xspace}
\usepackage{bm}
\usepackage{fancyref}
\usepackage{textcomp}

\usepackage{multirow}
\usepackage{stmaryrd}

\newenvironment{textbmatrix}{	\setlength{\arraycolsep}{2.5pt}%
								\big[\begin{matrix}}{\end{matrix}\big]%
								\raisebox{0.08ex}{\vphantom{M}}}

\def\be{\begin{equation}}
\def\ee{\end{equation}}
\def\een{\nonumber \end{equation}}
\def\mat{\begin{bmatrix}}
\def\emat{\end{bmatrix}}
\def\btm{\begin{textbmatrix}}
\def\etm{\end{textbmatrix}}

\def\ba#1\ea{\begin{align}#1\end{align}}
\def\bas#1\eas{\begin{align*}#1\end{align*}}
\def\bs#1\es{\begin{split}#1\end{split}} 
\def\bg#1\eg{\begin{gather}#1\end{gather}}
\def\bml#1\eml{\begin{multline}#1\end{multline}}
\def\bi#1\ei{\begin{itemize}#1\end{itemize}}

\safemath{\dirac}{\delta}					
\safemath{\krond}{\dirac}					

\safemath{\upto}{\uparrow}
\safemath{\downto}{\downarrow}
\safemath{\iu}{j}							
\safemath{\ev}{\lambda}						
\safemath{\hilseqspace}{l^{2}}				
\newcommand{\banachfunspace}[1]{\setL^{#1}}	
\safemath{\hilfunspace}{\banachfunspace{2}}	

\safemath{\SNR}{\text{\sc snr}} 				
\safemath{\No}{N_0}							
\safemath{\Es}{E_s}							
\safemath{\Eb}{E_b}							
\safemath{\EbNo}{\frac{\Eb}{\No}}
\safemath{\EsNo}{\frac{\Es}{\No}}

\DeclareMathOperator{\CHop}{\ensuremath{\opH}} 
\safemath{\tvir}{\rndh_{\CHop}}				
\safemath{\tvtf}{\rndl_{\CHop}}				
\safemath{\spf}{\rnds_{\CHop}}				
\safemath{\bff}{H_{\CHop}}					

\safemath{\ircf}{r_{h}}						
\safemath{\tftvcf}{r_{s}}					
\safemath{\tfcf}{r_{l}}						
\safemath{\bfcf}{r_{H}}						

\safemath{\tcorr}{c_h}						
\safemath{\scf}{c_{s}}						
\safemath{\tfcorr}{c_{l}}					
\safemath{\fcorr}{c_{H}}						

\safemath{\mi}{I}							
\safemath{\capacity}{C}						

\safemath{\normal}{\mathcal{N}}			
\safemath{\jpg}{\mathcal{CN}}			
\safemath{\mchain}{\leftrightarrow}		

\safemath{\dB}{\,\mathrm{dB}}
\safemath{\dBm}{\,\mathrm{dBm}}
\safemath{\Hz}{\,\mathrm{Hz}}
\safemath{\kHz}{\,\mathrm{kHz}}
\safemath{\MHz}{\,\mathrm{MHz}}
\safemath{\GHz}{\,\mathrm{GHz}}
\safemath{\s}{\,\mathrm{s}}
\safemath{\ms}{\,\mathrm{ms}}
\safemath{\mus}{\,\mathrm{\text{\textmu}s}}
\safemath{\ns}{\,\mathrm{ns}}
\safemath{\ps}{\,\mathrm{ps}}
\safemath{\meter}{\,\mathrm{m}}
\safemath{\mm}{\,\mathrm{mm}}
\safemath{\cm}{\,\mathrm{cm}}
\safemath{\m}{\,\mathrm{m}}
\safemath{\W}{\,\mathrm{W}}
\safemath{\mW}{\, \mathrm{mW}}
\safemath{\J}{\,\mathrm{J}}
\safemath{\K}{\,\mathrm{K}}
\safemath{\bit}{\,\mathrm{bit}}
\safemath{\nat}{\,\mathrm{nat}}

\safemath{\define}{\triangleq}			

\safemath{\equivalent}{\sim}
\safemath{\distas}{\sim}					
\safemath{\sdiff}{\Delta}				

\safemath{\reals}{\mathbb{R}}
\safemath{\positivereals}{\reals_{+}}
\safemath{\integers}{\mathbb{Z}}
\safemath{\posint}{\integers_{+}}
\safemath{\naturals}{\mathbb{N}}
\safemath{\posnaturals}{\naturals_{+}}
\safemath{\complexset}{\mathbb{C}}
\safemath{\rationals}{\mathbb{Q}}

\newcommand*{\fancyrefapplabelprefix}{app}		
\newcommand*{\fancyrefthmlabelprefix}{thm}		
\newcommand*{\fancyreflemlabelprefix}{lem}		
\newcommand*{\fancyrefcorlabelprefix}{cor}		
\newcommand*{\fancyrefdeflabelprefix}{def}		
\newcommand*{\fancyrefalglabelprefix}{alg}		
\newcommand*{\fancyrefproplabelprefix}{prop}		
\newcommand*{\fancyrefexmpllabelprefix}{exmpl}
\newcommand*{\fancyreftbllabelprefix}{tbl}
\frefformat{vario}{\fancyrefseclabelprefix}{Section~#1}
\frefformat{vario}{\fancyrefthmlabelprefix}{Theorem~#1}
\frefformat{vario}{\fancyreflemlabelprefix}{Lemma~#1}
\frefformat{vario}{\fancyrefcorlabelprefix}{Corrolary~#1}
\frefformat{vario}{\fancyrefdeflabelprefix}{Definition~#1}
\frefformat{vario}{\fancyrefalglabelprefix}{Algorithm~#1}
\frefformat{vario}{\fancyreffiglabelprefix}{Figure~#1}
\frefformat{vario}{\fancyrefapplabelprefix}{Appendix~#1} 
\frefformat{vario}{\fancyrefeqlabelprefix}{(#1)}
\frefformat{vario}{\fancyrefproplabelprefix}{Proposition~#1}
\frefformat{vario}{\fancyrefexmpllabelprefix}{Example~#1}
\frefformat{vario}{\fancyreftbllabelprefix}{Table~#1}

\safemath{\dictab}{[\,\dicta\,\,\dictb\,]}

\safemath{\ysig}{\bmy}
\safemath{\ysighat}{\hat{\ysig}}
\safemath{\ysigdim}{M}
\safemath{\xsig}{\bmx}
\safemath{\xsigdim}{N}
\safemath{\nx}{n_x}
\safemath{\zsig}{\bmz}
\safemath{\zsigdim}{\ysigdim}
\safemath{\rsig}{\bmr}
\safemath{\Adict}{\bA}
\safemath{\Adicttilde}{\widetilde{\Adict}}
\safemath{\Adictdim}{\outputdim\times\xsigdim}
\safemath{\avec}{\bma}
\safemath{\avectilde}{\tilde{\avec}}
\safemath{\Bdict}{\bB}
\safemath{\Bdicttilde}{\widetilde{\Bdict}}
\safemath{\Cdict}{\bC}
\safemath{\cvec}{\bmc}
\safemath{\Ddict}{\bD}
\safemath{\Ddictdim}{\ysigdim\times\xsigdim}
\safemath{\dvec}{\bmd}
\safemath{\Ddicttilde}{\widetilde{\bD}}
\safemath{\Bonb}{\bB}
\safemath{\bvec}{\bmb}
\safemath{\Bonbdim}{\ysigdim\times\ysigdim}
\safemath{\noise}{\bmn}
\safemath{\noisedim}{\ysigim}
\safemath{\err}{\bme}
\safemath{\errdim}{\ysigdim}
\safemath{\errset}{\setE}
\safemath{\nerr}{n_e}
\safemath{\delop}{\bP_\errset}
\safemath{\delopc}{\bP_{{\errset}^c}}

\safemath{\cplxi}{\imath}
\safemath{\cplxj}{\jmath}

\safemath{\dict}{\matD}
\safemath{\inputdim}{N}		
\safemath{\outputdim}{M}		
\safemath{\sparsity}{S}	
\safemath{\inputdimA}{{N_a}}	
\safemath{\inputdimB}{{N_b}}	
\safemath{\elemA}{{n_a}}	
\safemath{\elemB}{{n_b}}	
\safemath{\resA}{\matR_a}	
\safemath{\resB}{\matR_b}	
\safemath{\subD}{\matS} 
\safemath{\subA}{\matS_a} 
\safemath{\subB}{\matS_b} 
\safemath{\dicta}{\matA} 	
\safemath{\dictb}{\matB} 	
\safemath{\hollowS}{H}
\safemath{\hollowA}{H_a}
\safemath{\hollowB}{H_b}
\safemath{\cross}{Z}
\safemath{\coh}{\mu_d}			
\safemath{\coha}{\mu_a}			
\safemath{\cohb}{\mu_b}			
\safemath{\mubs}{\nu}	
\safemath{\cohm}{\mu_m} 
\safemath{\dictset}{\setD}	
\safemath{\dictsetp}{\dictset(\coh,\coha,\cohb)}	
\safemath{\dictsetgen}{\dictset_\text{gen}}
\safemath{\dictsetgenp}{\dictsetgen(\coh)}
\safemath{\dictsetonb}{\dictset_\text{onb}}
\safemath{\dictsetonbp}{\dictsetonb(\coh)}

\safemath{\leftside}{U}
\safemath{\rightsideA}{R_a}
\safemath{\rightsideB}{R_b}

\safemath{\indexS}{\setI_S} 

\safemath{\na}{n_a}			
\safemath{\nb}{n_b}			
\safemath{\coeffa}{p_i}	
\safemath{\coeffb}{q_j}	
\safemath{\seta}{\setP}		
\safemath{\setb}{\setQ}     
\safemath{\setw}{\setW}	
\safemath{\setz}{\setZ}	
\safemath{\cola}{\veca}		
\safemath{\colb}{\vecb}		
\safemath{\cold}{\vecd}		
\safemath{\inputvec}{\vecx} 	
\safemath{\error}{\vece}	
\safemath{\noiseout}{\vecz} 	
\safemath{\inputvecel}{x}
\safemath{\inputveca}{\vecx_a}
\safemath{\inputvecb}{\vecx_b}
\safemath{\outputvec}{\vecy}	
\safemath{\lambdamin}{\lambda_{\mathrm{min}}}

\safemath{\elltwo}{\ell_2}
\safemath{\ellone}{\ell_1}
\safemath{\ellzero}{\ell_0}
\safemath{\ellinf}{\ell_\infty}
\safemath{\licard}{Z(\coh,\coha,\cohb)}
\safemath{\xsol}{\hat{x}}
\safemath{\xbord}{x_b}		
\safemath{\xstat}{x_s}		
\safemath{\xstatLone}{\tilde{x}_s}
\safemath{\order}{\mathcal{O}} 
\safemath{\scales}{\Theta} 
\safemath{\ones}{\mathbf{1}} 
\safemath{\zeroes}{\mathbf{0}} 
\safemath{\thlone}{\kappa(\coh,\cohb)} 
\safemath{\constoneA}{\delta} 
\safemath{\constoneB}{\epsilon} 
\safemath{\nlarge}{L}				   
\safemath{\sumlarge}{S_\nlarge}
\safemath{\maxlarger}{P_\nlarge}	   
\safemath{\Pzero}{\textrm{P0}}	
\safemath{\Pone}{\textrm{P1}}
\safemath{\vecfir}{\vecw}			 
\safemath{\vecsec}{\vecz}
\safemath{\elvecfir}{w}              
\safemath{\elvecsec}{z}				 
\safemath{\nlargefir}{n}
\safemath{\normout}{\gamma}
\safemath{\auxfun}{h}
\safemath{\supp}{\textrm{supp}}

\safemath{\indexa}{\ell}
\safemath{\indexb}{r}
\safemath{\indexc}{i}
\safemath{\indexd}{j}

\safemath{\project}{P}

\begin{document}

\title{Data-Mining Textual Responses \\ to Uncover Misconception Patterns}
%
%
%
%
%

\numberofauthors{3} 
%
\author{
%
%
\alignauthor
Joshua J. Michalenko\\
       \affaddr{Rice University}\\
       \email{jjm7@rice.edu}
\alignauthor
Andrew S. Lan \\
       \affaddr{Princeton University}\\
       \email{andrew.lan@princeton.edu}
\alignauthor Richard G. Baraniuk\\
       \affaddr{Rice University}\\
       \email{richb@rice.edu}
}

\maketitle
\begin{abstract}
An important, yet largely unstudied, problem in student data analysis is to detect \emph{misconceptions} from students' responses to \emph{open-response} questions.  Misconception detection enables instructors to deliver more targeted feedback on the misconceptions exhibited by many students in their class, thus improving the quality of instruction.  In this paper, we propose a new natural language processing-based framework to detect the common misconceptions among students' textual responses to short-answer questions.  We propose a probabilistic model for students' textual responses involving misconceptions and experimentally validate it on a real-world student-response dataset.  Experimental results show that our proposed framework excels at classifying whether a response exhibits one or more misconceptions.  More importantly, it can also automatically detect the common misconceptions exhibited across responses from multiple students to multiple questions;  this property is especially important at large scale, since instructors will no longer need to manually specify all possible misconceptions that students might exhibit.  
\end{abstract}



\section{Introduction}

\sloppy
The rapid developments of large-scale learning platforms (e.g., MOOCs (edx.org, coursera.org) and OpenStax Tutor (openstaxtutor.org)) have enabled not only access to high-quality learning resources to a large number of students, but also the collection of student data at very large scale.  The scale of this data presents a great opportunity to revolutionize education by using machine learning algorithms to \emph{automatically} deliver personalized analytics and feedback to students and instructors in order to improve the quality of teaching and learning.

\fussy

\subsection{Detecting misconceptions from\\ student-response data} 

The predominant form of student data, their \emph{responses} to assessment questions, contain rich information on their knowledge.
Analyzing why a student answers a question incorrectly is of crucial importance to deliver timely and effective feedback.  
Among the possible causes for a student to answer a question incorrectly, exhibiting one or more \emph{misconceptions} is critical, since upon detection of a misconception, it is very important to provide targeted feedback to a student to correct their misconception in a timely manner.  
Examples of using misconceptions to improve teaching include incorporating misconceptions to design better distractors for multiple-choice questions \cite{pavliklick}, implementing a dialogue-based tutor to detect misconceptions and provide corresponding feedback to help students self-practice \cite{dialoguetutor}, preparing prospective instructors by examining the causes of common misconceptions among students \cite{teachertrain}, and incorporating misconceptions into item response theory (IRT) for learning analytics \cite{tatsuokarule1}.  

The conventional way of leveraging misconceptions is to rely on a set of pre-defined misconceptions provided by domain experts \cite{grade12chem,pavliklick,teachertrain,dialoguetutor}.  
However, this approach is not scalable, since it requires a large amount of human effort and is domain-specific.  
With the large scale of student data at our disposal, a more scalable approach is to automatically detect misconceptions from data.  

Recently, researchers have developed approaches for data-driven misconception detection; most of these approaches analyze students' response to \emph{multiple-choice} questions.  
Examples of these approaches include detecting misconceptions in multiple-choice mathematics questions and modeling students' progress in correcting them \cite{kenlick} via the additive factor model \cite{afm}, and clustering students' responses across a number of multiple-choice physics questions \cite{april}.  
However, multiple-choice questions have been shown to be inferior to open-response questions in terms of pedagogical value \cite{ofvsmc}.
Indeed, students' responses to open-response questions can offer deeper insights into their knowledge state.

To date, detecting misconceptions from students' responses to open-response questions has largely remained an unexplored problem.  
A few recent developments work exclusively with \emph{structured} responses, e.g., sketches \cite{sketchopenformmc}, short mathematical expressions \cite{mathopenformmc}, group discussions in a chemistry class \cite{schmidtchem}, and algebra with simple syntax \cite{elmadanidatadriven}.

\subsection{Contributions}

In this paper, we propose a natural language processing framework that detects students' common misconceptions from their \emph{textual} responses to open-response, short-answer questions.  
This problem is very difficult, since the responses are, in general, \emph{unstructured}. 

Our proposed framework consists of the following steps.  First, we transform students' textual responses to a number of short-answer questions into low-dimensional textual feature vectors using several well-known word-vector embeddings.  
These tools include the popular Word2Vec embedding \cite{word2vec}, the GLOVE embedding \cite{pennington2014glove}, and an embedding based on the long-short term memory (LSTM) neural network \cite{Palangi:2016,SHochreiter1997}. 
We then propose a new statistical model that jointly models both the transformed response textual feature vectors and expert labels on whether a response exhibits one or more misconceptions;  these labels identify only \emph{whether or not} a response exhibits one or more misconceptions but not \emph{which} misconception it exhibits.  

Our model uses a series of latent variables: the feature vectors corresponding to the correct response to each question, the feature vectors corresponding to each misconception, the tendency of each student to exhibit each misconception, and the confusion level of each question on each misconception.  
We develop a Markov chain Monte Carlo (MCMC) algorithm for parameter inference under the proposed statistical model.
We experimentally validate the proposed framework on a real-world educational dataset collected from high school classes on AP biology.  

Our experimental results show that the proposed framework excels at classifying whether a response exhibits one or more misconceptions compared to standard classification algorithms and significantly outperforms a baseline random forest classifier.  We also compare the prediction performance across all three embeddings.  
More importantly, we show examples of common misconceptions detected from our dataset and discuss how this information can be used to deliver targeted feedback to help students correct their misconceptions.

\section{Dataset and pre-processing}

In this section, we first detail our short-answer response dataset, and then detail our pre-processing approach to convert responses into vectors using word-to-vector embeddings.

\subsection{Dataset}

Our dataset consists of students' textual responses to short-answer questions in high school classes on AP Biology administered on OpenStax Tutor \cite{ost}.  Every response was labeled by an expert grader as to whether it exhibited one or more misconceptions.  A total of $N = 386$ students each responded to a subset of a total of $Q = 1668$ questions; each response was manually labeled by one or multiple expert graders, resulting in a total of $\sim 60,000$ labeled responses.  Since there is no clear rubric defining what is a misconception, graders might not necessarily agree on what label to assign to each response.  Therefore, we trim the dataset to only keep responses that are labeled by multiple graders and they also assigned the same label, resulting in $13,099$ responses.  We also further trim the dataset by filtering out students who respond to less than 5 questions and questions with less than 5 responses in every dataset.  This subset contains $6,152$ responses.  

The questions in our dataset are drawn from the OpenStax AP biology textbook; we divide the full dataset into smaller subsets corresponding to each of the first four units \cite{osbio}, since different units correspond to entirely different sub-areas in biology.  These units cover the following topics:
\begin{itemize}
\item{Unit 1: The Chemistry of Life, Chapters 1-3}
\item{Unit 2: The Cell, Chapters 4-10}
\item{Unit 3: Genetics, Chapters 11-17}
\item{Unit 4: Evolutionary Processes, Chapters 18-20}
\end{itemize}
To summarize, we show the dimensions of the subsets of the data corresponding to each unit in Table~\ref{tbl:dataStat}.  Since not every student was assigned to every question, the dataset is sparsely populated;  Table~\ref{tbl:dataStat} also shows the portion of responses that are observed in the trimmed data subsets, denoted as ``sparsity''.

\begin{table}[tp]
	\centering
	\caption{Statistics of subsets of the AP Biology datasets that correspond to each unit.} \label{tbl:dataStat}
	\vspace{0.2cm}
	\scalebox{1.00}{ 
		\begin{tabular}{cccc}
		\toprule
		&  $N$ & $Q$ & Sparsity (\%)\\
		\midrule
		Unit 1 & 47 & 77 & 0.280\\
		\midrule
		Unit 2 & 101 & 104 & 0.243\\
		\midrule
		Unit 3 & 73 & 91 & 0.236\\
		\midrule
		Unit 4 & 43 & 75 & 0.315\\
		\bottomrule
		\end{tabular}
	}
\end{table}

\subsection{Response embeddings}

We first perform a pre-processing step by transforming each textual student response into a corresponding real-valued vector via three different word-vector embeddings.  
Our first embedding uses the Word2Vec embedding \cite{word2vec} trained on the OpenStax Biology textbook (an approach also mentioned in \cite{openformtext}), to learn embeddings that put more emphasis on the technical vocabulary specific to each subject. 
We create the feature vector for each response by mapping each individual word in the response to its corresponding feature vector, and then adding them together.  
Concretely, denote  $\mathbf{x}_{i,j} = \{w_1, w_2,...,w_{T_{i,j}} \}$ as the collection of words in the textual response of student~$j$ to question~$i$, where $T_{i,j}$ denotes the total number of words in this response (excluding common stopwords).  
We then map each word $w_t$ to its corresponding $D$-dimensional feature vector $r(w_t) \in \mathbb{R}^D$ using the trained Word2Vec model.  We use $D = 10$ for the Word2Vec embedding. We then compute the student response feature vector as $\mathbf{f}_{i,j} = \sum_{t=1}^{T_{i,j}} r(w_t)$.

Our second word-vector embedding is a pre-trained GLOVE embedding with $D=25$ \cite{pennington2014glove}.  The GLOVE embedding is very similar to the Word2Vec embedding, with the main difference being that it takes corpus-level word co-occurrence statistics into account.  Moreover, the quality of the GLOVE embedding for common words is likely higher since it is pre-trained on a huge corpus (comparing to only the OpenStax Biology textbook for Word2Vec).

Both the Word2Vec embedding and the GLOVE embedding do not take word ordering into account, and for misconception classification, this drawback can lead to problems.  For example, responses ``If X then Y'' and ``If Y then X'' may have completely different meanings depending on the context, where it's possible for one to exhibit a common misconception while the other one does not.  Using the Word2Vec and GLOVE embeddings, these responses will be embedded to the same feature vector $\mathbf{f}_{i,j}$, making them indistinguishable from each other.  Therefore, our third word-vector embedding is based on the long short-term memory (LSTM) neural network, which is a recurrent neural network that excels at capturing long-term dependencies in sequential data.  Therefore, it can take word ordering into account, a feature that we believe is critical for misconception detection.  We implement a 2-layer LSTM network with 10 hidden units and train it on the OpenStax Biology textbook.  For each student response, we use the text as character-by-character inputs to the LSTM network and use the last layer's hidden unit activation values (stacked in a $D=10$ dimensional vector) as its textual feature $\mathbf{f}_{i,j}$.

\section{Statistical Model}

\begin{figure}[t]
\vspace{0.0cm}
\centering
\includegraphics[width=\columnwidth]{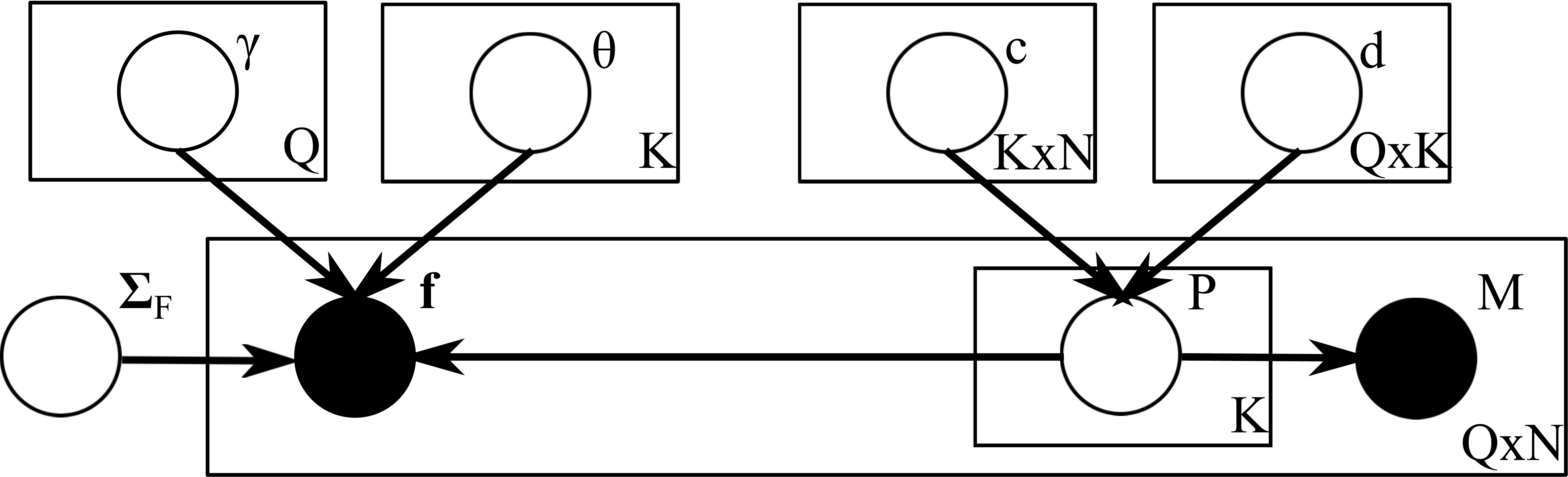}
\vspace{-0.0cm}
\caption{Visualization of the statistical model.  Black nodes denote observed data; white nodes denote latent variables to be inferred.} \label{fig:plate}
\vspace{-0.0cm}
\end{figure}

We now detail our statistical model; its graphical model is visualized in Figure~\ref{fig:plate}. 
Concretely, let there be a total of $N$ students, $Q$ questions, and $K$ misconceptions.  
Let $M_{i,j} \in \{0,1\}$ denote the binary-valued misconception label on the response of student~$j$ to question~$i$ provided by an expert grader, with $j \in \{ 1, \ldots, N\}$ and $i \in \{ 1, \ldots, Q\}$, where $1$ represents the presence of (one or more) misconceptions, and $0$ represents no misconceptions.

We transform the raw text of student~$j$'s response to question~$i$ into a $D$-dimensional real-valued feature vector, denoted by $\vecf_{i,j} \in \mathbb{R}^D$, via a pre-processing step (detailed in the previous section).  
Let $\Omega \subseteq \{ 1, \ldots, Q\} \times \{ 1, \ldots, N\}$ denote the subset of student responses that are labeled, since every student only responds to a subset of the questions.

We denote the \emph{tendency} of student~$j$ to exhibit misconception~$k$, with $k \in \{1,\ldots,K\}$ as $c_{k,j} \in \mathbb{R}$, and the \emph{confusion level} of question~$i$ on misconception~$k$, as $d_{i,k} \in \mathbb{R}$. 
Then, let $P_{i,j,k} \in \{0,1\}$ denote the binary-valued latent variable that represents whether student~$j$ exhibits misconception~$k$ in their response to question~$i$, with $1$ denoting that the misconception is present and $0$ otherwise.  
We model $P_{i,j,k}$ as a Bernoulli random variable 
\begin{align} \label{eq:p}
p(P_{i,j,k} = 1) = \Phi(c_{k,j} + d_{i,k}),  \quad  (i,j) \in \Omega,
\end{align}
where $\Phi(x) = \int_{-\infty}^x \mathcal{N}(t;0,1) \mathrm{d}t$ denotes the inverse probit link function (the cumulative distribution function of the standard normal random variable). 
Given $P_{i,j,k} \; \forall k$, we model the observed misconception label $M_{i,j}$ as 
\begin{align} \label{eq:m}
M_{i,j} = \left \{ \begin{array}{ll}
0 &\text{if} \; P_{i,j,k} = 0  \;\, \forall k, \\[0.1cm]
1 & \text{otherwise}, 
\end{array} \right. \quad (i,j) \in \Omega.
\end{align}
In words, a response is labeled as having a misconception if one or more misconceptions is present (given by the latent misconception exhibition variables $P_{i,j,k}$). 
Given $P_{i,j,k} \; \forall k$, the textual response feature vector that corresponds to student~$j$'s response to question~$i$, $\vecf_{i,j}$, is modeled as 
\begin{align} \label{eq:f}
\vecf_{i,j} \distas \mathcal{N}(\boldsymbol{\gamma}_i + \sum_k P_{i,j,k} \boldsymbol{\theta}_k, \boldsymbol{\Sigma}_F), \quad \forall (i,j) \in \Omega,
\end{align}
where $\boldsymbol{\gamma}_i$ denotes the feature vector that corresponds to the correct response to question~$i$, $\boldsymbol{\theta}_k$ denotes the feature vector that corresponds to misconception~$k$, and $\boldsymbol{\Sigma}_F$ denotes the covariance matrix of the multivariate normal distribution characterizing the feature vectors.  
In other words, the feature vector of each response is a \emph{mixture} of the feature vectors corresponding to the correct response to the question and each misconception the student exhibits.  

In the next section, we develop an MCMC inference algorithm to infer the values of the latent variables $\boldsymbol{\gamma}_i$, $\boldsymbol{\theta}_k$, $\boldsymbol{\Sigma}_F$, $P_{i,j,k}$, $c_{k,j}$, and $d_{i,k}$, given observed data $\vecf_{i,j}$ and $M_{i,j}$.

\section{Parameter Inference}
\label{sec:algo}

We use a Gibbs sampling algorithm \cite{gelmanbook} for parameter inference under the proposed statistical model.  The prior distributions of the latent variables are listed as follows:
\begin{align*}
\boldsymbol{\gamma}_i & \distas \mathcal{N}(\boldsymbol{\mu}_\gamma, \boldsymbol{\Sigma}_\gamma), \\
\boldsymbol{\theta}_k & \distas \mathcal{N}(\boldsymbol{\mu}_\theta, \boldsymbol{\Sigma}_\theta), \\
\boldsymbol{\Sigma}_F & \distas IW(h_F, \bV_F), \\
c_{k,j} & \distas \mathcal{N}(\mu_c,\sigma_c^2), \\
d_{i,k} & \distas \mathcal{N}(\mu_d,\sigma_d^2),
\end{align*}
where $IW(\cdot)$ denotes the inverse-Wishart distribution and $\boldsymbol{\mu}_\gamma$, $\boldsymbol{\Sigma_\gamma}$, $\boldsymbol{\mu}_\theta$, $\boldsymbol{\Sigma_\theta}$, $h_F$, $\bV_F$, $\mu_c$, $\sigma_c^2$, $\mu_d$, and $\sigma_d^2$ are hyperparameters. 


We start by randomly initializing the values of the latent variables $\boldsymbol{\gamma}_i$, $\boldsymbol{\theta}_k$, $\boldsymbol{\Sigma}_F$, $P_{i,j,k}$, $c_{k,j}$, $d_{i,k}$, $a_j$, and $\mu_i$ by sampling from their prior distributions.  Then, in each iteration of our Gibbs sampling algorithm, we iteratively sample the value of each random variable from its full conditional posterior distribution.  Specifically, in each iteration, we perform the following steps:
\begin{itemize}
\item[a) Sample $P_{i,j,k}$:] We first sample the latent misconception indicator variable $P_{i,j,k}$ from its posterior distribution as
\begin{align*}
P_{i,j,k} \! = \! \left \{ \!\!\! \begin{array}{cl}
0 &\text{if} \; M_{i,j} = 0, \\[0.1cm]
1 & \text{if} \; M_{i,j} = 1 \; \text{and} \; P_{i,j,k'} = 0 \; \forall \; k' \neq k, \\
\frac{r}{r+1} & \text{if} \; M_{i,j}  = 1 \; \text{and} \; \exists \; k' \neq k \; \text{s.t.} \; P_{i,j,k'} \!\! = \!\! 1,
\end{array} \right.
\end{align*}
where
\begin{align*}
r & = \frac{p(\vecf_{i,j} | \boldsymbol{\gamma}_i, \boldsymbol{\theta}_k, \forall k, \boldsymbol{\Sigma}_F, P_{i,j,k' \neq k}, P_{i,j,k} = 1)}{p(\vecf_{i,j} | \boldsymbol{\gamma}_i, \boldsymbol{\theta}_k, \forall k, \boldsymbol{\Sigma}_F, P_{i,j,k' \neq k}, P_{i,j,k} = 0)} \cdot \\
& \quad \quad \frac{p(P_{i,j,k} = 1 | c_{k,j}, d_{i,k} )}{p(P_{i,j,k} = 0 | c_{k,j}, d_{i,k})}.
\end{align*}
The terms in the expression above are given by \fref{eq:p} and \fref{eq:f}.

\item[b) Sample $\boldsymbol{\gamma}_i$:] We then sample the feature vector that corresponds to the correct response to each question, $\boldsymbol{\gamma}_i$, from its posterior distribution as $\boldsymbol{\gamma}_i \! \distas \! \mathcal{N}(\boldsymbol{\mu}_{\gamma_i},\boldsymbol{\Sigma}_{\gamma_i})$ where
\begin{align*}
\boldsymbol{\mu}_{\gamma_i} & = \boldsymbol{\Sigma}_{\gamma_i} \!\!\! \left(\boldsymbol{\Sigma}_\gamma^{-1} \boldsymbol{\mu}_\gamma + \boldsymbol{\Sigma}_F^{-1} \!\!\!\!\!\! \sum_{j: (i,j) \in \Omega} \!\!\!\!\!\! (\vecf_{i,j} - \sum_k P_{i,j,k} \boldsymbol{\theta}_k)\right),\\
\boldsymbol{\Sigma}_{\gamma_i} & = (\boldsymbol{\Sigma}_\gamma^{-1} + n_i \boldsymbol{\Sigma}_F^{-1})^{-1},
\end{align*}
where $n_i = \sum_j I\left((i,j) \in \Omega \right)$. 

\item[c) Sample $\boldsymbol{\theta}_k$:] We then sample the feature vector that corresponds to each misconception, $\boldsymbol{\theta}_k$, from its posterior distribution as $\boldsymbol{\theta}_k \distas \mathcal{N}(\boldsymbol{\mu}_{\theta_k},\boldsymbol{\Sigma}_{\theta_k})$ where
\begin{align*}
\boldsymbol{\mu}_{\theta_k} & \! = \! \boldsymbol{\Sigma}_{\theta_k} \!\!\! \left( \!\! \boldsymbol{\Sigma}_\theta^{-1} \boldsymbol{\mu}_\theta \! + \! \boldsymbol{\Sigma}_F^{-1} \!\!\!\!\!\!\!\! \sum_{i,j: P_{i,j,k} = 1} \!\!\!\!\!\!\! (\vecf_{i,j} \!- \! \boldsymbol{\gamma}_i \! - \!\! \sum_{k' \neq k} \!\! P_{i,j,k'} \boldsymbol{\theta}_{k'}) \!\! \right) \!\!,\\
\boldsymbol{\Sigma}_{\theta_k} &  = (\boldsymbol{\Sigma}_\theta^{-1} + n_k \boldsymbol{\Sigma}_F^{-1})^{-1},
\end{align*}
where $n_k = \sum_{i,j} I\left( P_{i,j,k} = 1 \right)$. 

\item[d) Sample $\boldsymbol{\Sigma}_F$:] We then sample the covariance matrix $\boldsymbol{\Sigma}_F$ from its posterior distribution as
\begin{align*}
\boldsymbol{\Sigma}_F & \distas IW \left(h_F + n, \bV_F + \bM) \right),
\end{align*}
where $n  \!\! = \!\! \sum_{i,j} I \left((i,j) \in \Omega\right)$ and $\bM = \sum_{i,j: (i,j) \in \Omega} (\vecf_{i,j} - \boldsymbol{\gamma}_i - \sum_k P_{i,j,k} \boldsymbol{\theta}_k) (\vecf_{i,j} - \boldsymbol{\gamma}_i - \sum_k P_{i,j,k} \boldsymbol{\theta}_k)^T$. 

\item[e) Sample $c_{k,j}$ and $d_{i,k}$:] In order to sample $c_{k,j}$ and $d_{i,k}$, we first sample the value of the auxiliary variable $z_{i,j,k}$ (following the standard approach proposed in \cite{albertchib}) as 
\begin{align*}
z_{i,j,k} \distas \mathcal{N}^\pm(c_{k,j} + d_{i,k}, 1), \forall (i,j) \in \Omega,
\end{align*}
where $\mathcal{N}^\pm(\cdot)$ denotes the truncated normal random distribution truncated to the positive side when $P_{i,j,k} = 1$ and negative side when $P_{i,j,k} = 0$.  We then sample $c_{k,j}$ from its posterior distribution as
\begin{align*}
c_{k,j} \distas \mathcal{N}(\mu_{c_{k,j}},\sigma_{c_{k,j}}^2),
\end{align*}
where $n_j = \sum_i I\left((i,j) \in \Omega \right)$, $\sigma_{c_{k,j}}^2 = 1/(1/\sigma_c^2 + n_j)$, and $\mu_{c_{k,j}} \!\! = \!\!\sigma_{c_{k,j}}^2 \!\! \left(\mu_c /\sigma_c^2 + \sum_{i: (i,j) \in \Omega} (z_{i,j,k} - d_{i,k}) \right)$.  We then sample $d_{i,k}$ from its posterior distribution as
\begin{align*}
d_{i,k} \distas \mathcal{N}(\mu_{d_{i,k}},\sigma_{d_{i,k}}^2),
\end{align*}
where $\sigma_{d_{i,k}}^2 \!\! = \!\! 1/(1/\sigma_d^2 + n_i)$, and $\mu_{d_{i,k}} = \sigma_{d_{i,k}}^2 (\mu_d /\sigma_d^2 + \sum_{j: (i,j) \in \Omega} (z_{i,j,k} - c_{k,j}) )$.
\end{itemize}

We run the iterations detailed above for a number of $T$ total iterations with a certain burn-in period, and use the samples of each latent variable to approximate their posterior distributions.

\sloppy
\paragraph{Label switching}  Parameter inference under our model suffers from the label-switching issue that is common in mixture models \cite{gelmanbook}, meaning that the mixture components might be permuted between iterations.  We employ a post-processing step to resolve this issue.  Specifically, we first calculate the augmented data likelihood at each iteration (indexed by $\ell$) as
\begin{align*}
L^\ell & = \prod_{i,j} p(\vecf_{i,j} \!\mid\! \boldsymbol{\gamma}_i^\ell, P_{i,j,k}^\ell, \boldsymbol{\theta}_k^\ell, \forall k) \prod_{i,j,k} p(P_{i,j,k}^\ell \!\mid\! c_{k,j}^\ell, d_{i,k}^\ell) \\
& = \prod_{i,j} \mathcal{N}(\vecf_{i,j} \!\mid\! \boldsymbol{\gamma}_i^\ell + \sum_k P_{i,j,k}^\ell \boldsymbol{\theta}_k^\ell, \boldsymbol{\Sigma}_F^\ell) \times \\
& \quad \prod_{i,j,k} \Phi( (2 P_{i,j,k}^\ell -1) (c_{k,j}^\ell + d_{i,k}^\ell)).
\end{align*}
Then, we identify the iteration $\ell_\text{max}$ with the largest augmented data likelihood, and permute the variables $\boldsymbol{\theta}_k^\ell$, $c_{k,j}^\ell$, and $d_{i,k}^\ell$ that best match the variables $\boldsymbol{\theta}_k^{\ell_\text{max}}$, $c_{k,j}^{\ell_\text{max}}$, and $d_{i,k}^{\ell_\text{max}}$.  After this post-processing step, we can simply calculate the posterior means of each one of these sets of variables by taking averages of their values across non burn-in iterations.

\begin{figure*}[t]
    \centering
    \subfigure[Unit 1 ]{\includegraphics[scale=0.095]{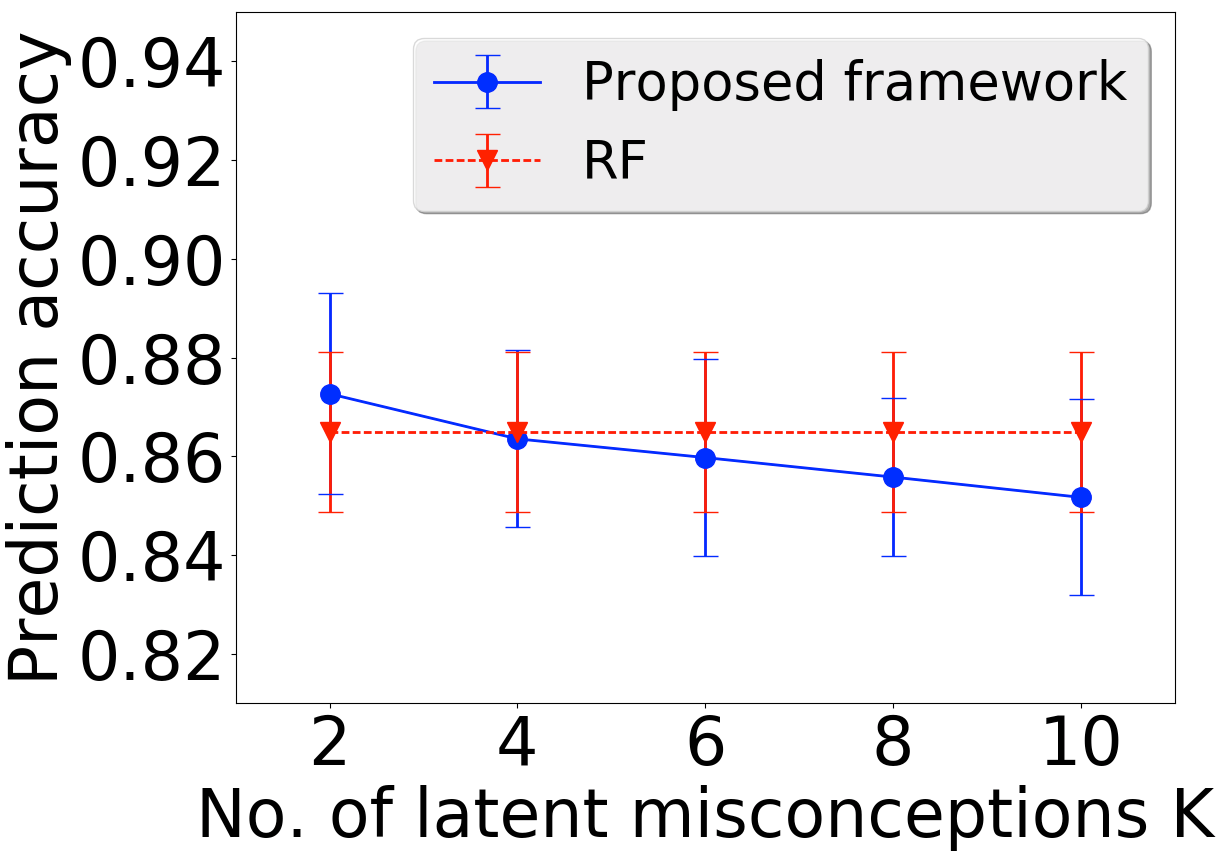}}
    \hspace{0cm}
    \subfigure[Unit 2 ]{\includegraphics[scale=0.095]{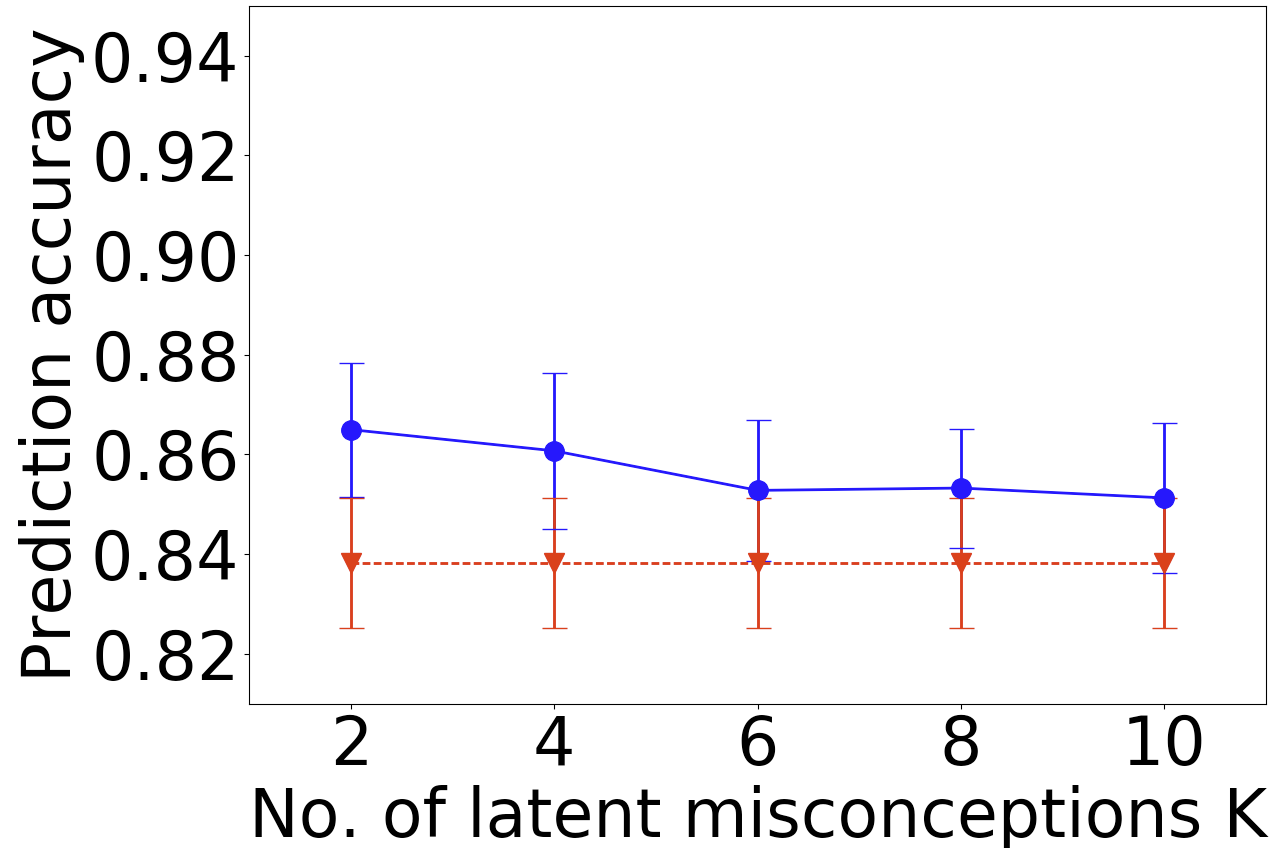}}
    \hspace{0cm}
    \subfigure[Unit 3 ]{\includegraphics[scale=0.095]{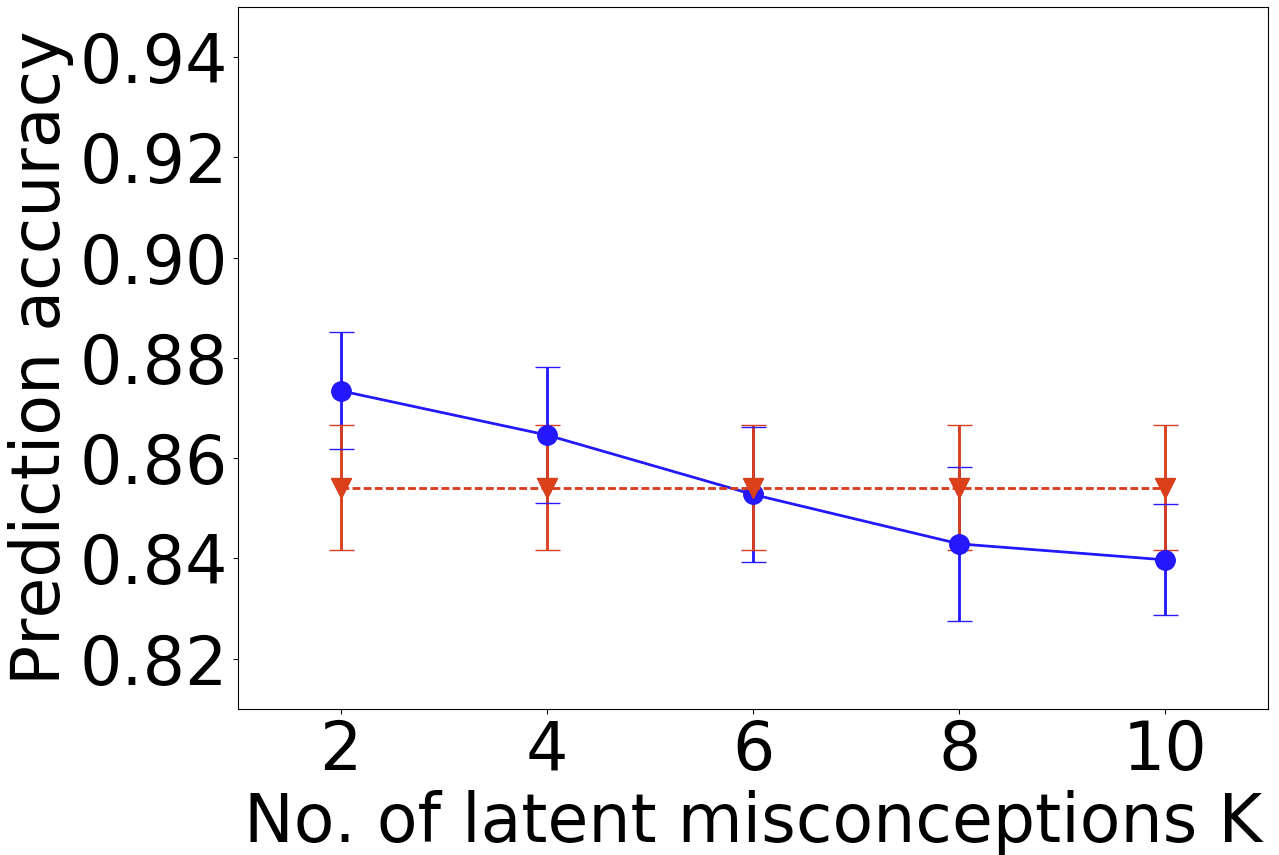}}
    \hspace{0cm}
    \subfigure[Unit 4 ]{\includegraphics[scale=0.095]{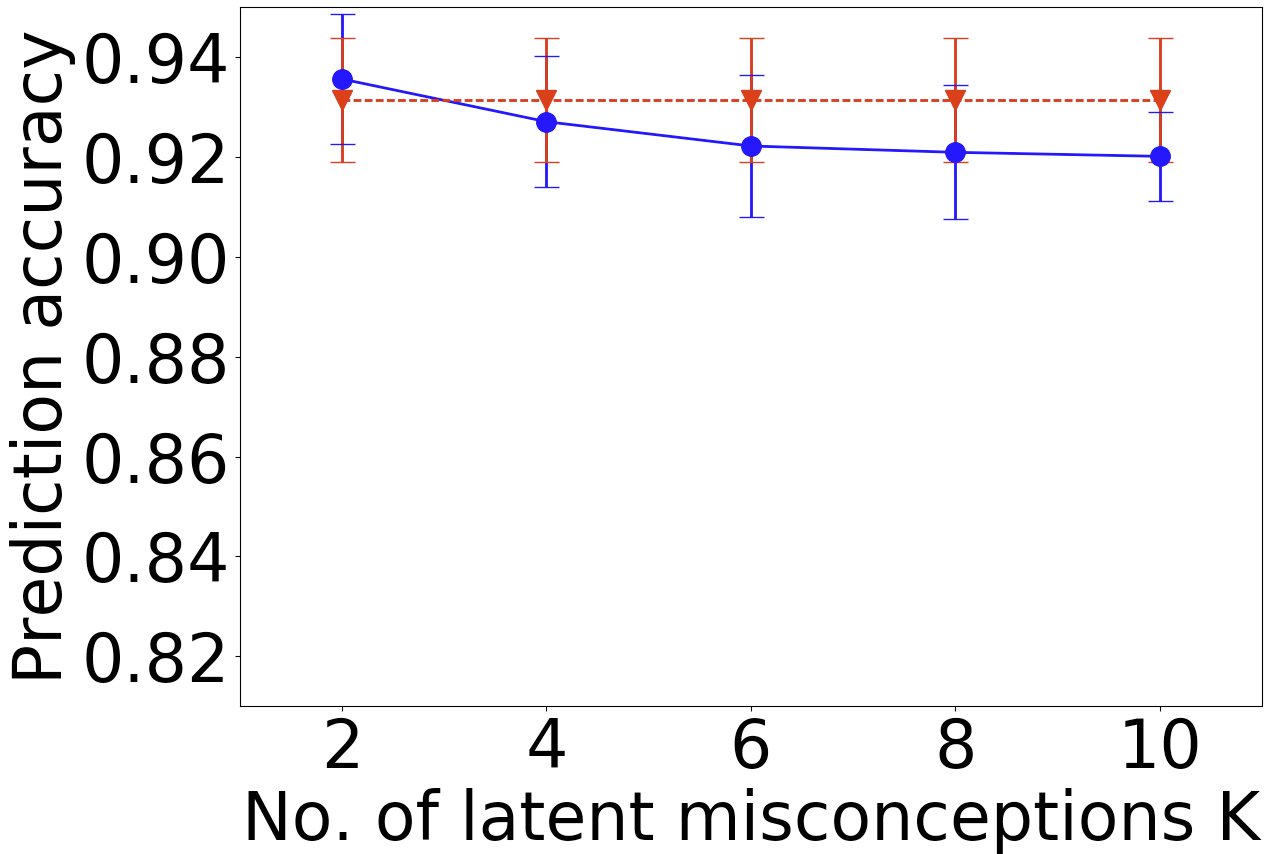}}
    \caption{Comparison of the prediction performance of the proposed model against RF on our AP Biology dataset using the ACC metric as the number of latent misconceptions $K$ varies, with the LSTM embedding.}
    \label{fig:D_LSTM10_accuracy}
    \end{figure*}

\fussy
\section{Experiments}

We experimentally validate the efficacy of the proposed framework using our AP Biology class dataset.  We first compare the proposed framework against a baseline random forest (RF) classifier that classifies whether a student response exhibits one or more misconceptions.  We then show common misconceptions detected in our datasets and discuss how the proposed framework can use this information to deliver meaningful targeted feedback to students that helps them correct their misconceptions. 

\subsection{Experimental setup}

We run our experiments with $K \in \{ 2,4,6,8,10 \} $ latent misconceptions with hyperparameters $\boldsymbol{\mu}_\gamma = \boldsymbol{\mu}_\theta = \mathbf{0}_D$, $\boldsymbol{\Sigma}_\gamma = \boldsymbol{\Sigma}_\gamma = \bV_F =  \bI_D$, $h_F = 10$, $\mu_c = \mu_d = 0$, and $\sigma_c^2 = \sigma_d^2 = 1$, for a total of $T = 500$ iterations with the first $250$ iterations as burn-in.  We compare the proposed framework against a baseline random forest (RF) classifier\footnote{The RF classifier achieves the best performance among a number of off-the-shelf baseline classifiers, e.g., logistic regression, support vector machines, etc. Therefore, we do not compare it against other baseline classifiers.} using the textual response feature vectors $\vecf_{i,j}$ to classify the binary-valued misconception label $M_{i,j}$, with 200 decision trees.

We randomly partition each dataset into 5 folds and use 4 folds as the training set and the other fold as the test set.  We then train the proposed framework and RF on the training set and evaluate their performance on the test set, using two metrics: i) prediction accuracy (ACC), i.e., the portion of correct predictions, and ii) area under curve (AUC), i.e., the area under the receiver operating characteristic (ROC) curve of the resulting binary classifier \cite{accauc}.  
Both metrics take values in $[0,1]$, with larger values corresponding to better prediction performance. 
We repeat our experiments for $20$ random partitions of the folds.  

For the proposed framework, the predictive probability that a response with its feature vector $\vecf_{i,j}$ exhibits a misconception, i.e., the probability that at least one of the $K$ latent misconception exhibition state variables take the value of $1$, is given by $1 - \widehat{p}_{i,j}$, where
\begin{align*}
\widehat{p}_{i,j} & = p(M_{i,j} = 0 \!\mid\! \vecf_{i,j}, \boldsymbol{\gamma}_i, \boldsymbol{\Sigma}_F, \boldsymbol{\theta}_k, \forall k, c_{k,j}, d_{i,k}) \\
& =  \frac{p(\vecf_{i,j} \!\!\mid\!\! \boldsymbol{\theta}_k, P_{i,j,k} = 0, \forall k) \prod_k p(P_{i,j,k} = 0 \!\!\mid\!\! c_{k,j}, d_{i,k})}{\sum_{P_{i,j,k}, \, \forall k} \! (p(\vecf_{i,j} \!\mid\! \boldsymbol{\theta}_k, P_{i,j,k} \forall k) \prod_k p(P_{i,j,k} \!\!\mid\!\! c_{k,j}, d_{i,k}))},
\end{align*}
where in the last expression we omitted the conditional dependency of $\vecf_{i,j}$ on $\boldsymbol{\gamma}_i$ and $\boldsymbol{\Sigma}_F$ due to spatial constraints.  For RF, the predictive probability is given by the fraction of decision trees that classifies $M_{i,j} = 1$ given $\vecf_{i,j}$.

\subsection{Results and discussions}

\sloppy
The number of latent misconceptions $K$ is an important parameter controlling the granularity of the misconceptions that we aim to detect.  Figure~\ref{fig:D_LSTM10_accuracy} shows the comparison between the proposed framework using different values of $K$ and RF using the ACC metric with the LSTM embedding.  We see an obvious trend that, as $K$ increases, the prediction performance decreases.  The likely cause of this trend is that the proposed framework tends to overfit as the number of latent misconceptions grows very large since some of our datasets do not contain very rich misconception types.  
Moreover, the number of common misconceptions varies across different units, with Unit~$2$ likely containing more misconception types than Units~$1$ and $4$. 

\fussy

We then compare the performance of the proposed framework against RF on misconception label classification in \fref{tbl:word2vec} using $K=2$ and all three embeddings.  Tables~\ref{tbl:word2vec}-\ref{tbl:lstm} show comparisons of the proposed framework against RF using both the ACC and AUC metrics on all three different word embeddings.
The proposed framework significantly outperforms RF (1--4\% using the ACC metric and 4-18\% using the AUC metric) on almost all 4 data subsets using every embedding.  
The only case where the proposed framework does not outperform RF is on Unit~$1$ using the GLOVE embedding.  We postulate that the reason for this result is that this unit is about chemistry and has a lot of responses with more chemical molecular expressions than words; therefore, the proposed framework does not have enough textual information to exhibit its advantages (grouping responses that share the same misconceptions into clusters) over the simple classifier RF.

Both the proposed framework and RF perform much better using the GLOVE and LSTM embeddings than the Word2Vec embedding.  This result is likely due to the fact that these embeddings are more advanced than the Word2Vec embedding: the GLOVE embedding considers additional word co-occurrence statistics than the Word2Vec embedding, is trained on a much larger corpus, and has a higher dimension $D=25$, while the LSTM embedding is the only embedding that takes word ordering into account.  Moreover, both algorithms perform best on Unit~$4$, which is likely due to two reasons: i) the Unit~$4$ subset has a larger portion of its responses labeled, and ii) Unit~$4$ is about evolution, which results in responses that are much longer and thus contains richer textual information.

\begin{table*}[t]
\centering
\caption{Performance comparison on misconception label classification of a textual response in terms of the prediction accuracy (ACC) and area under the receiver operating characteristic curve (AUC) of the proposed framework against a random forest (RF) classifier, using the AP Biology dataset and the Word2Vec embedding.} \label{tbl:word2vec}
\vspace{-0.0cm}
\scalebox{0.8}{ 
\begin{tabular}{ccccccccc}
\toprule
 & \multicolumn{2}{c}{Unit 1} & \multicolumn{2}{c}{Unit 2} & \multicolumn{2}{c}{Unit 3} & \multicolumn{2}{c}{Unit 4} \\
\cmidrule(l){2-3} \cmidrule(l){4-5} \cmidrule(l){6-7} \cmidrule(l){8-9}
& ACC & AUC & ACC & AUC & ACC & AUC & ACC & AUC \\
\midrule
Proposed framework & ${\bf 0.789\!\pm\! 0.014}$ & ${\bf 0.762 \!\pm\! 0.027}$ & ${\bf 0.774 \!\pm\! 0.015}$ & ${\bf 0.758 \!\pm\! 0.023}$ & ${\bf 0.779 \!\pm\! 0.019}$ & ${\bf 0.752 \!\pm\! 0.020}$ & $ {\bf0.887 \!\pm\! 0.011}$ & ${\bf 0.774 \!\pm\! 0.029}$ \\
\midrule
RF & $0.762\!\pm\! 0.019$ & $0.645\!\pm\! 0.025$ & $0.735\!\pm\! 0.011$ & $0.676\!\pm\! 0.014$ & $0.758\!\pm\! 0.017$ & $0.630\!\pm\! 0.024$ & $0.873\!\pm\! 0.009$ & $0.604\!\pm\! 0.034$  \\
\bottomrule
\end{tabular}
}
\end{table*}

\begin{table*}
\centering
\caption{Performance comparison on misconception label classification of a textual response in terms of the prediction accuracy (ACC) and area under the receiver operating characteristic curve (AUC) of the proposed framework against a random forest (RF) classifier, using the AP Biology dataset and the GLOVE embedding.} \label{tbl:glove}
\vspace{-0.0cm}
\scalebox{0.8}{ 
\begin{tabular}{ccccccccc}
\toprule
 & \multicolumn{2}{c}{Unit 1} & \multicolumn{2}{c}{Unit 2} & \multicolumn{2}{c}{Unit 3} & \multicolumn{2}{c}{Unit 4} \\
\cmidrule(l){2-3} \cmidrule(l){4-5} \cmidrule(l){6-7} \cmidrule(l){8-9}
& ACC & AUC & ACC & AUC & ACC & AUC & ACC & AUC \\
\midrule
Proposed framework & $0.867 \!\pm\! 0.014$ & ${\bf 0.762 \!\pm\! 0.048}$ & ${\bf 0.870 \!\pm\! 0.010}$ & ${\bf 0.821 \!\pm\! 0.024}$ & ${\bf 0.893 \!\pm\! 0.017}$ & ${\bf 0.794 \!\pm\! 0.039}$ & ${\bf 0.953 \!\pm\! 0.015}$ & ${\bf 0.892 \!\pm\! 0.047}$ \\
\midrule
RF & ${\bf 0.876\!\pm\! 0.014}$ & $0.697\!\pm\! 0.022$ & $0.859\!\pm\! 0.013$ & $0.771\!\pm\! 0.040$ & $0.883\!\pm\! 0.008$ & $0.616\!\pm\! 0.043$ & $0.948\!\pm\! 0.019$ & $0.731\!\pm\! 0.006$  \\
\bottomrule
\end{tabular}
}
\end{table*}

\begin{table*}
\centering
\caption{Performance comparison on misconception label classification of a textual response in terms of the prediction accuracy (ACC) and area under the receiver operating characteristic curve (AUC) of the proposed framework against a random forest (RF) classifier, using the AP Biology dataset and the LSTM embedding.} \label{tbl:lstm}
\vspace{-0.0cm}
\scalebox{0.8}{ 
\begin{tabular}{ccccccccc}
\toprule
 & \multicolumn{2}{c}{Unit 1} & \multicolumn{2}{c}{Unit 2} & \multicolumn{2}{c}{Unit 3} & \multicolumn{2}{c}{Unit 4} \\
\cmidrule(l){2-3} \cmidrule(l){4-5} \cmidrule(l){6-7} \cmidrule(l){8-9}
& ACC & AUC & ACC & AUC & ACC & AUC & ACC & AUC \\
\midrule
Proposed framework &${\bf 0.873 \!\pm\! 0.042}$ & ${\bf 0.772 \!\pm\! 0.093}$ & ${\bf 0.865 \!\pm\! 0.025}$ & ${\bf 0.829 \!\pm\! 0.044}$ & ${\bf 0.873 \!\pm\! 0.027}$ & ${\bf 0.792 \!\pm\! 0.061}$ & ${\bf 0.936 \!\pm\! 0.032}$ & ${\bf 0.832 \!\pm\! 0.094}$ \\
\midrule
RF & ${ 0.865\!\pm\! 0.035}$ & $0.711\!\pm\! 0.086$ & $0.838\!\pm\! 0.028$ & $0.722\!\pm\! 0.043$ & $0.854\!\pm\! 0.028$ & $0.697\!\pm\! 0.057$ & $0.931\!\pm\! 0.025$ & $0.709\!\pm\! 0.105$  \\
\bottomrule
\end{tabular}
}
\end{table*}

\subsection{Uncovering common misconceptions}

We emphasize that, in addition to the proposed framework's significant improvement over RF in terms of misconception label classification, it features great interpretability since it identifies common misconceptions from data.  As an illustrative example, the following responses from multiple students across two questions are identified to exhibit the same misconception in the Unit~$4$ subset using the Word2Vec embedding:

\medskip
\begin{mdframed}
{\em Question~1}: People who breed domesticated animals try to avoid inbreeding even though most domesticated animals are indiscriminate. Evaluate why this is a good practice.\\
{\em Correct Response}: A breeder would not allow close relatives to mate, because inbreeding can bring together deleterious recessive mutations that can cause abnormalities and susceptibility to disease. \\
{\bf Student Response~1}: Inbreeding can cause a rise in unfavorable or detrimental traits such as genes that cause individuals to be prone to disease or have unfavorable mutations. \\
{\bf Student Response~2}: Interbreeding can lead to harmful mutations.
\end{mdframed}
\begin{mdframed}
{\em Question~2}: When closely related individuals mate with each other, or inbreed, the offspring are often not as fit as the offspring of two unrelated individuals. Why? \\
{\em Correct Response}: Inbreeding can bring together rare, deleterious mutations that lead to harmful phenotypes. \\
{\bf Student Response~3}: Leads to more homozygous recessive genes thus leading to mutation or disease.\\
{\bf Student Response~4}: When related individuals mate it can lead to harmful mutations.
\end{mdframed}
\medskip

Although these responses are from different students to different questions, they exhibit one common misconception, that inbreeding leads to harmful mutations.  
Once this misconception is identified, course instructors can deliver the targeted feedback that inbreeding only brings together harmful mutations, leading to issues like abnormalities, rather than directly leading to harmful mutations.

Moreover, the proposed framework can automatically discover common misconceptions that students exhibit without input from domain experts, especially when the number of students and questions are very large.  Specifically, in the example above, we are able to detect such a common misconception that 4 responses exhibit by analyzing the 1016 responses in the AP Biology Unit 4 dataset; however, it would not likely be detected if the number of responses was smaller and fewer students exhibited the misconception.  This feature makes it an attractive data-driven aid to domain experts in designing educational content to address student misconceptions. 

We show another example that the proposed framework can automatically group student responses to the same group according to the misconceptions they exhibit.  The example shows two detected common misconceptions among students' responses to a single question in the Unit~$2$ subset using the LSTM embedding:

\medskip
\begin{mdframed}
{\em Question}: What is the primary energy source for cells? \\
{\em Correct response}: Glucose. \\
{\bf Student responses with misconception~$1$:}\\
sunlight \\
sun \\
The sun \\
he sun? \\
{\bf Student responses with misconception~$2$:}\\
ATP \\
adenosine triphosphate \\
ATPPPPPPPPPPPPP \\
atp mitochondria
\end{mdframed}
\medskip

We see that the proposed framework has successfully identified two common misconception groups, with incorrect responses that list ``sun'' and ``ATP'' as the primary energy source for cells.  Note that the LSTM embedding enables it to assign the full and abbreviated form of the same entity (``adenosine triphosphate'' and ``ATP'') into the same misconception cluster, without employing any pre-processing on the raw textual response data.  The likely reason for this result is that our LSTM embedding is trained on a character-by-character level on the OpenStax Biology textbook, where these terms appear together frequently, thus enabling the LSTM to transform them into similar vectors.  This observation highlights the importance of using good, information-preserving word-vector embeddings for the proposed framework to maximize its capability of detecting common misconceptions. 

\section{Conclusions and Future Work}

In this paper, we have proposed a natural language processing-based framework for detecting and classifying common misconceptions in students' textual responses.  
Our proposed framework first transforms their textual responses into low-dimensional feature vectors using three existing word-vector embedding techniques, and then estimates the feature vectors characterizing each misconception, among other latent variables, using a proposed mixture model that leverages information provided by expert human graders.  Our experiments on a real-world educational dataset consisting of students' textual responses to short-answer questions showed that the proposed framework excels at classifying whether a response exhibits one or more misconceptions.  Our proposed framework is also able to group responses with the same misconceptions into clusters, enabling the data-driven discovery of common misconceptions without input from domain experts. 

Possible avenues of future work include i) automatically generate the appropriate feedback to correct each misconception, ii) leverage additional information, such as the text of the correct response to each question, to further improve the performance on predicting misconception labels, iii) explore the relationship between the dimension of the word-vector embeddings and prediction performance, and iv) develop embeddings for other types of responses, e.g., mathematical expressions \cite{mlp} and chemical equations. 

\newpage
\balance

\bibliographystyle{abbrv}
\bibliography{sparfaclustbib}

\end{document}